# Bayesian Stratified Sampling to Assess Corpus Utility

Judith Hochberg, Clint Scovel, Timothy Thomas, and Sam Hall
Group CIC-3, Los Alamos National Laboratory
Mail Stop B265
Los Alamos, NM 87545
{judithh, jcs, trt, shall}@lanl.gov

## Abstract

This paper describes a method for asking statistical questions about a large text corpus. We exemplify the method by addressing the question, "What percentage of *Federal Register* documents are real documents, of possible interest to a text researcher or analyst?" We estimate an answer to this question by evaluating 200 documents selected from a corpus of 45,820 *Federal Register* documents. Stratified sampling is used to reduce the sampling uncertainty of the estimate from over 3100 documents to fewer than 1000. The stratification is based on observed characteristics of real documents, while the sampling procedure incorporates a Bayesian version of Neyman allocation. A possible application of the method is to establish baseline statistics used to estimate recall rates for information retrieval systems.

## Introduction

The traditional task in information retrieval is to find documents from a large corpus that are relevant to a query. In this paper we address a related task: answering *statistical* questions about a corpus. Instead of finding the documents that match a query, we quantify the percentage of documents that match it.

The method is designed to address statistical questions that are:

- *subjective*: that is, informed readers may disagree about which documents match the query, and the same reader may make different judgment at different times. This characteristic describes most queries of real interest to text researchers.

- *difficult*: that is, one cannot define an algorithm to reliably assess individual documents, and thus the corpus as a whole. This characteristic follows naturally from the first. It may be compounded by an insufficient understanding of a corpus, or a shortcoming in one's tools for analyzing it.

Statistical questions asked of small corpora can be answered exhaustively, by reading and scoring every document in the corpus. Such answers will be subjective, since judgments about the individual documents are subjective. For a large corpus, it is not feasible to read every document. Instead, one must sample a subset of documents, then extrapolate the results of the sample to the corpus as a whole. The conclusions that one draws from such a sampling will have two components: the estimated answer to the question, and a confidence interval around the estimate.

The method described in this paper combines traditional statistical sampling techniques (Cochran (1963), Kalton (1983)) with Bayesian analysis (Bayes (1763), Berger (1980)) to reduce this sampling uncertainty. The method is well-grounded in statistical theory, but its application to textual queries is novel. One begins by stratifying the data using objective tests designed to yield relatively homogeneous strata, within which most documents either match or do not match the query. Then one samples randomly within each stratum, with the number of documents sampled per stratum determined through the analysis of a presample. A reader scores each selected document, and the results of the different strata are combined. If the strata are well constructed, the resulting estimate about the corpus will have a much smaller *credibility interval* (the Bayesian version of a confidence interval) than one based on a sample of the corpus as a whole.

The method is well suited for subjective queries because it brings a human reader's subjective judgments to bear on individual documents. The Bayesian approach that we apply to this problem allows a second opportunity for the reader to influence the results of the sampling. The reader can construct a probability density that summarizes his or her prior expectations about each stratum. These prior expectations are combined with pre-sampling results to determine the makeup of the final sample. When the final sample is analyzed, the prior expectations are again factored in, influencing the estimated mean and the size of the credibility interval. Thus different readers' prior expectations, and their judgments of individual documents, can lead to substantially different results, which is consistent with the subjective probability paradigm.

In earlier work we used this method to analyze medical records, asking, "What percentage of the patients are female?" (Thomas et al. (1995)). The lack of a required gender field in the record format made this a subjective question, especially for records that did not specify the patient's gender at all, or gave conflicting clues. We stratified the corpus into probable male and female records based on linguistic tests such as the number of female versus male pronouns in a record, then sampled within each stratum. Stratification reduced the sampling uncertainty for the question from fourteen percentage points (based on an overall sample of 200 records) to five (based on a stratified sample of the same size).

In this paper, we update the method and apply it to a new corpus, the *Federal Register*. The main change from Thomas et al. (1995) is a greater focus on numerical methods as opposed to parametric and formulaic calculations. For example, we use a non-parametric prior density instead of a beta density, and combine posterior densities between strata using a Monte Carlo simulation rather than weighted means and variances. Other differences, such as a Bayesian technique for allocating samples between strata, and a new method for determining the size of the credibility interval, are noted in the text.

The *Federal Register* corpus is of general interest because it is part of the TIPSTER collection. The question we addressed is likewise of general interest: what percentage of documents are of possible interest to a researcher, or to an analyst querying the corpus? Anyone who has worked with large text corpora will recognize that not all documents are created equal; identifying and filtering uninteresting documents can be a nuisance. Estimating the percentage of uninteresting documents in a corpus therefore helps determine its utility.

The paper begins by describing the *Federal Register* corpus and the corpus utility query. It then describes two steps in finding a statistical answer to the query: first through an overall sample of 200 documents from the corpus, then through a stratified sample of 200, then 400 documents. The Conclusion takes up the question of possible application domains and implementation issues for the method.

## 1 Data

The text corpus used in this study was the *Federal Register*. Published by the United States Government, the *Register* contains the full text of all proposed and final Federal rules and regulations, notices of meetings and programs, and executive proclamations. We used an electronic version of the *Register* that was part of the 1997 TIPSTER collection distributed by the Linguistic Data Consortium (http://www.ldc.upenn.edu/). It consisted of 348 files, each purported to contain one issue of the *Register* for the years 1988 and 1989. Each separate rule, regulation, etc. within an issue was considered a separate document and was bracketed with SGML markup tags <DOC> and </DOC>. The corpus contained 45,820 such documents.

There were systematic differences between the corpus and the printed version of the *Federal Register*. The on-line version omitted page numbers and boilerplate text seen in the printed version. The order of documents in the two versions differed; for example, documents within special Parts following the main body of the printed version were intermixed with the main body of the on-line version. Other differences, such as missing or repeated documents, were less systematic and appeared to be errors.

Thus the TIPSTER corpus could in no way be considered a perfect electronic version of the *Federal Register*. Rather, it should be considered a realistic example of archival records that are not extensively edited for the purposes of information extraction research.

## 2 The Query

The query we addressed in this paper grew out of an attempt to establish basic statistics for *Federal Register* documents. When counting documents and determining their length, we noticed that some purported documents (as judged by <DOC> </DOC> bracketing) were not what we came to define as **real** *Federal Register* documents: documents describing the activities of the federal government. Besides real documents, the electronic *Register* contained pseudo-documents related to the use and publication of the paper version of the *Register*, such as tables of contents, indices, blank pages, and title pages.

This discovery at first appeared to be a mere nuisance. We assumed that there was an easy way to separate pseudo-documents from real documents, but could not find one. The harder we looked for a way to separate the two document types, the more we realized that this distinction had theoretical interest. Determining the percentage of real documents would serve to evaluate the true size of the corpus, and its usefulness for TIPSTER type applications where documents relevant to topic queries are expected to be returned.

This query matched the two criteria set forth in the Introduction for applicability to our method. As described above, there was no easy way to separate real documents from pseudo-documents. The query was also subjective, since readers might disagree about the classification of particular documents. For example, a document announcing classes on how to use the *Federal Register* could be considered a real document (since notices of all sorts appear in the *Register*), or a pseudo-document (since it is promulgated by the *Register*'s office and appears at regular intervals). As another example, readers might disagree about which erratum documents are significant enough to be considered real documents themselves.

## 3  Overall estimation and the Bayesian approach

We will illustrate the Bayesian approach in the context of a straw man effort to estimate the percentage of real documents without stratifying the corpus. From the entire 45,820 document set, we sampled 200 documents at random. Sampling was done with replacement (i.e., we did not remove sampled documents from the population); however, no documents were observed to be selected twice. One of our researchers then reviewed the documents and judged them as real documents versus pseudo-documents. He did this by reading the first fifty lines of each document.

Of the 200 documents sampled, 187, or 0.935, were judged to be real documents; this served as our initial estimate for the overall percentage of real documents in the corpus. We then used Bayesian techniques to modify this estimate based on our prior expectations about the population. This was a three-step process. First, we calculated the binomial likelihood function corresponding to the sampling results. Second, we encoded our prior expectations in a likelihood function. Third, we combined the binomial and prior likelihood functions to create a *posterior probability density*. This posterior served as the basis for the final estimate and credibility interval.

### 3.1  Binomial likelihood function

The standard binomial likelihood function associated with the sampling result (187 real documents out of 200),

$$f(x) = \frac{200!}{187!13!} \, x^{187} \, (1-x)^{13}, \qquad (1)$$

is graphed in Figure 1. It shows, given each possible true percentage of real documents, the likelihood that one would find 187 real documents out of 200 sampled. We evaluated the likelihood function at a high degree of granularity -- at x intervals corresponding to five significant digits -- so that we would later be able to map percentages of documents onto exact numbers of documents.

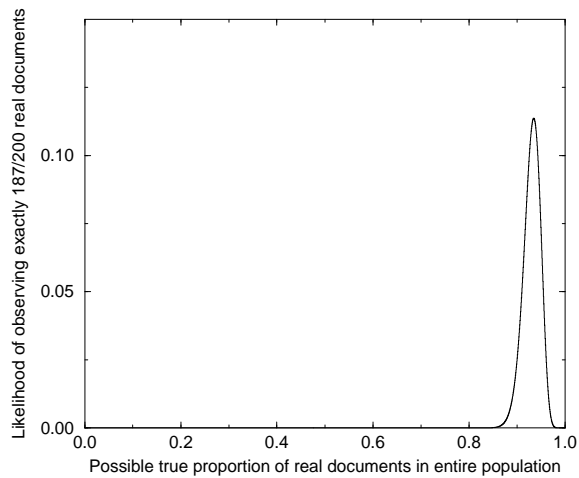

Figure 1.  Binomial likelihood function given 187 real documents out of 200 sampled

### 3.2  Prior

We chose a prior by inputting a personal likelihood: one researcher's subjective opinion about the population based on a first look at the corpus. The researcher's input consisted of eleven likelihood values, at intervals of 0.1 on the x axis, as shown in Figure 2. These points were then splined to obtain a likelihood function (Fig. 2; see Press et al. (1988)) and normalized to obtain a probability density. The resulting density was discretized at five significant digits to match the granularity of the binomial likelihood function.

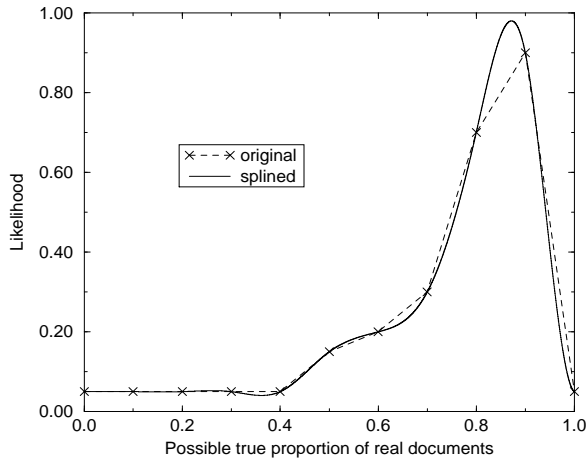

Figure 2. Prior personal likelihood for proportion of real documents

An alternative to the above procedure is to choose a prior from a parametric family such as beta densities. This approach simplifies later calculations, as shown in Thomas et al. (1995). However, the non-parametric prior allows the researcher more freedom to choose a probability density that expresses his or her best understanding of a population.

### 3.3 Posterior

Once the prior density was established, we applied Bayes' theorem to calculate a *posterior probability density* for the population. We did this by multiplying binomial likelihood function (Fig. 1) by the prior density (Fig. 2), then normalizing. The non-zero portion of the resulting posterior is graphed in Figure 3.

Figure 3 contrasts this posterior density with the binomial likelihood function from Figure 1, also normalized. From a Bayesian perspective, the latter density implicitly factors in the standard *non-informative prior* in which each possible percentage of real documents has an equal probability. The informative prior shifted the density slightly to the left.

We used the posterior density to revise our estimate of the percentage of real documents in the population, and to quantify the uncertainty of this estimate. The revised estimate was the mean $\mu$, of the density, defined as $\sum_{k=1}^{l} x_k f(x_k)$, where $l$ is the number of points evaluated for the function (1,000,001). This evaluated to 0.9257. To quantify the uncertainty of this estimate, we found the 95% credibility interval surrounding it -- that is, the range on the x axis that contained 95% of the area under the posterior density.

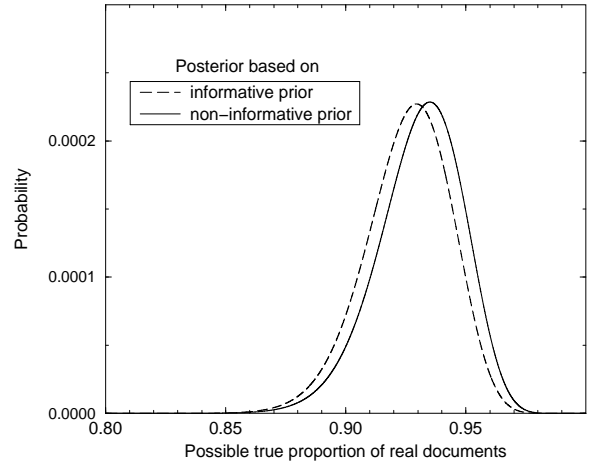

Figure 3. Posterior probability density for proportion of real documents

The traditional way to find this interval is to assume a normal distribution, compute the variance $\sigma^2$ of the posterior, defined as $\sum_{k=1}^{l} f(x_k)(x_k-\mu)^2$, and set the credibility interval at $\mu \pm 1.96 \sigma$. This yielded a credibility interval between 0.8908 and 0.9606. As an alternative, we calculated the credibility interval exactly, in a numerical fashion that yielded the tightest possible interval and thus somewhat reduced the final uncertainty of the estimate. To do so we moved outward from the peak of the density, summing under the curve until we reached a total probability of 0.95. At each step outwards from the peak we considered probability values to the right and left and chose the larger of the two. This method also finds a tighter interval than the numerical method used in Thomas et al. (1995), which was based on finding the left and right tails that each contained 0.025 of the density.

The credibility interval found for the posterior probability density using the exact method is summarized in Table 1, in percentages of real documents and in numbers of real documents. The document range was calculated by multiplying the percentage range by the number of documents in the corpus (45,820). For comparison's sake the table includes the parallel results obtained using a

| Table 1. Results from overall sampling (200 documents) ||||
| Posterior based on which prior | Interval || Size of credibility interval (in documents) |
| --- | --- | --- | --- |
| | In percent real documents | In number of documents | |
| Informative | $0.89029 < p < 0.95902$ | 40793-43942 | 3149 |
| Non-informative | $0.89519 < p < 0.96374$ | 41017-44158 | 3141 |

non-informative prior. The two intervals were almost identical. The non-informative prior led to a slightly smaller credibility interval than the informative prior, implying that the latter was poorly chosen. But regardless of the prior used, the size of the credibility interval, expressed in numbers of documents, was over 3100 documents. This was a lot of uncertainty -- enough to taint any decision about the usage of documents in the on-line *Federal Register*.

## 4 Reducing uncertainty using stratified sampling

We performed a stratified sampling to reduce the uncertainty displayed in Table 1. This process involved dividing the data into two relatively homogeneous strata, one containing mostly real documents, the other mostly pseudo-documents, and combining sampling results from the two strata.

This approach is advantageous because the variance of a binomial density, $\frac{p(1-p)}{n}$ (where $n$ is the number sampled, and $p$ the percentage of "yes" answers), shrinks dramatically for extreme values of $p$. Therefore, one can generally reduce sampling uncertainty by combining results from several homogeneous strata, rather than doing an overall sample from a heterogeneous population.

As with our overall sample, we performed the stratified sampling within the Bayesian framework. The steps described in Section 3 for the overall sample were repeated for each stratum (with an additional step to allocate samples to the strata), and the posteriors from the strata were combined for the final estimate.

### 4.1 Defining strata and allocating the samples

We divided the documents into two strata: apparent real documents, and apparent pseudo-documents. The basis for the division was the observation that most pseudo-documents were of the following types:

1. Part dividers (title pages for subparts of an issue)
2. Title pages
3. Tables of contents
4. Reader Aids sections
5. Instructions to insert illustrations not present in the electronic version
6. Null documents (no text material between <TEXT> and </TEXT> markers)
7. Other defective documents, such as titles of presidential proclamations that were separated from the proclamation itself.

We wrote a short Perl script that recognized pseudo-document types 1-4 using key phrases (e.g., />Part [IVXM]/ for Part dividers), and types 5-7 by their short length. This test stratified the data into 3444 apparent pseudo-documents and 42,376 apparent real documents.

Exploration of the strata showed that this stratification was not perfect -- indeed, if it were, we could no longer call this query difficult! Some real documents were misclassified as pseudo-documents because they accidentally triggered the key phrase detectors. An erratum document correcting the incomprehensible *Register*-ese error

"<ITAG tagnum=68>BILLING CODE 1505-01-D </ITAG>"

was misclassified as a real document. However, we will see that the stratification sufficed to sharply reduce the credibility interval.

Before doing the stratified sampling, we had to decide how many documents to sample from each stratum. In a departure from Thomas et al. (1995), we used a Bayesian modification of Neyman allocation to do this. Traditional Neyman allocation requires a pre-sampling of each stratum to determine its heterogeneity; heterogeneous strata are then sampled more intensively. In Newbold's Bayesian modification (1971), prior expectations for each stratum are combined with pre-sample results to create a posterior density for each stratum. These posteriors are then used to determine the allocation.

This technique therefore required creating posterior densities for each stratum that blended a prior density and a presample. Accordingly, we devised priors for the two strata -- apparent pseudo-documents, and apparent real documents -- based on our exploratory analysis of the strata. As in the overall analysis (Section 3.2), we splined the priors to five significant digits. The original (unsplined) priors are graphed in Figure 4.

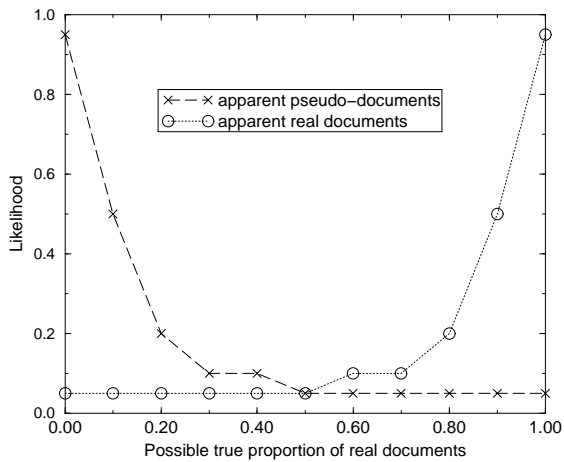

Figure 4. Prior likelihoods for proportion of real documents in the strata

For the presample, we randomly chose ten documents from each stratum (with replacement) and read and scored them. The presample results were perfect -- all apparent pseudo-documents were pseudo-documents, and all apparent real documents were real. We applied Bayes' theorem to calculate the posterior density for each stratum, multiplying the binomial likelihood function associated with the stratum's presample by the relevant prior density, and normalizing.

With these posteriors in hand, we were ready to determine the optimum allocation among the strata. Newbold (1971) gives the fraction $q_i$ allocated to each stratum $i$ by

$$q_i = \frac{C_i^{1/2} A_i^{1/2} (n_i+1)^{1/2}}{\sum_{j=1}^{k} C_j^{1/2} A_j^{1/2} (n_j+1)^{1/2}} \quad (2)$$

where $k$ is the number of strata, $C_i$ is the cost of sampling a stratum (assumed here to be 1), $n_i$ is the number of documents in the presample for the stratum, and $A_i$ is

$$A_i = \frac{\Pi_i^2 P_i (1-P_i)}{(n_i+2)} \quad (3)$$

where $\Pi_i$ is the fraction of the overall population that comes from the $i^{th}$ stratum, and $P_i$ is the population mean for the posterior density in the $i^{th}$ stratum. The outcome of this procedure was an allocation of 15 apparent pseudo-documents and 185 apparent real documents.

### 4.2 Posteriors for each stratum

Having already sampled ten documents from each stratum, we now sampled an additional 5 apparent pseudo-documents and 175 apparent real documents to make up the full sample. We chose documents randomly with replacement and judged each document subjectively as above. To our surprise (knowing that the stratification was error-prone), this sampling again gave perfect results: all apparent pseudo-documents were pseudo-documents, and all apparent real documents were real.

We applied Bayes' theorem a final time to derive a new posterior probability density for each stratum based on the results of the full sample. For each stratum, we multiplied the binomial likelihood function corresponding to the full sampling results (0/15 and 185/185) by the prior probability density for each stratum (i.e., the posterior density from the presample), then normalized.

### 4.3 Combining the results: Monte Carlo simulation

The final step was to combine the two posteriors to obtain an estimate and credibility interval for the population as a whole. The traditional approach would be to find the mean and variance for each stratum's posterior and combine these according to each stratum's weight in the population. Newbold (1971) gives the weighted mean as $\sum_{i=1}^{k} \Pi_i \frac{b_i}{n_i}$, where $b_i$ is the number of real documents found in stratum $i$ out of $n_i$ sampled. As an alternative technique, we used a Monte Carlo simulation (Shreider (1966)) to compute the density of the fraction of real documents $p = \sum_{i=1}^{k} \Pi_i p_i$. We then used this density to provide a final estimate and a corresponding credibility interval.

The Monte Carlo simulation combined the two posteriors in proportion to the incidence of real and pseudo-documents in the *Federal Register* corpus. Real documents constituted 0.925 of the corpus, and pseudo-documents the remaining 0.075. To perform the simulation, we randomly sampled both posterior densities a million times. For each pair of points picked, we determined the weighted average of the two points, and incremented the value of the corresponding point on the overall density by $10^{-6}$, or one millionth. For example, if we picked 0.2 from the posterior for apparent pseudo-documents and 0.9 from the posterior for apparent real documents, then we would increment the value of 0.8475 (0.2*0.075 + 0.9*0.925) in the overall density by $10^{-6}$. At the end of the simulation, the total area of the density was 1.0.

The resulting overall density is graphed in Figure 5 along with the posteriors. Since the corpus mostly contained apparent real documents, the combined density was closer to that stratum's density.

Using the same method as in section 3.3, we then found the exact 95% credibility interval for the combined density. The results, summarized in Table 2, show a better than 3:1 reduction from the overall sample, from 3138 to 919 documents. Table 2 also shows the results obtained using a non-informative prior -- that is, based on the sampled results alone, without any specific prior expectations. Here we clearly see the benefit of vigorously applying the Bayesian approach, as the prior knowledge helps reduce the credibility interval by seven-tenths of a percent, or 325 documents.

## Discussion and Conclusion

By sampling 200 documents, stratified according to observed document characteristics with a Bayesian version of Neyman allocation, we have addressed the question of how many *Federal Register* documents are useful documents that reflect the activities of the Federal government. The answer was a credibility interval between 91% and 93%, or between 41,768 and 42,687 documents. This was a substantially tighter estimate than could be obtained using either an overall sample, or a stratified sample without prior expectations.

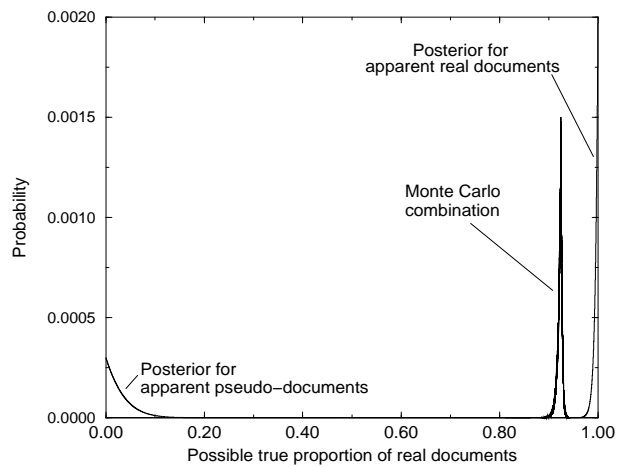

Figure 5. Posteriors after full sample, and Monte Carlo combination of posteriors

This estimate was probably tight enough to be useful in applications such as comparing the utility of different corpora. If higher precision were called for, the simplest way to further narrow the credibility interval would be to increase the sample size. In a follow-on experiment, it took less than a half hour to read an additional 200 documents (this turned up two incorrectly stratified documents, confirming our expectations from exploratory analysis). The new data sharpened the posteriors, reducing the combined credibility interval to 624 documents, or 1.3 percentage points. Further reductions could be obtained as desired.

A final topic to address is when and how our technique may be used. What types of questions are likely to be addressed, and what are the implementation issues involved?

| Table 2. Results from stratified sampling (200 documents) | | | |
|---|---|---|---|
| Posterior based on which prior | Interval | | Size of document interval |
| | In percent real documents | In number of documents | |
| Informative | 0.91157 < p < 0.93163 | 41,768-42,687 | 919 |
| Non-informative | 0.91074 < p < 0.93789 | 41,730-42,974 | 1244 |

We see two likely types of questions. A question may be asked for its own sake, as in this paper or Thomas et al. (1995). Looking further at the *Federal Register* corpus, other feasible questions using our method come to mind, such as:

- Has the amount of attention paid to the environment by the Federal government increased?
- What proportion of Federal affairs involve the state of New Mexico?

Users of other corpora could likewise pose questions relevant to their own interests.

A question could also be asked not for its own sake, but to establish a baseline statistic for information retrieval (IR) recall rates. Recall is the percentage of relevant documents for a query that an IR system actually finds. To establish recall, one must know how many relevant documents exist. The standard technique for estimating this number is "pooling": identifying relevant documents from among those returned by all IR systems involved in a comparison. This method is used by the TREC program (Voorhees and Harman (1997)). Our method is a principled alternative to this method that is well-grounded in statistical theory, and, unlike pooling, is independent of any biases present in current IR systems.

Applying the method to a new question, whether for its own sake or to determine recall, involves developing a stratification test, constructing a prior density for each stratum, performing the presample and full samples, and combining the results. Of these steps, stratification is the most important in reducing the credibility interval. In our work to date we have achieved good results with stratification tests that are conceptually and computationally simple. We suspect that when asking multiple questions of the same corpus, it may even be possible to automate the construction of stratification scripts. Priors are easiest to construct if the strata are clean and well-understood.

The appropriate amount of time to invest in refining a stratification test and the associated priors depends on the cost of evaluating documents and the importance of a small credibility interval. If documents are easy to evaluate, one might choose to put less time into stratification and priors construction, and reduce the credibility interval by increasing sample size. If one is restricted to a small sample, then accurate stratification and good priors are more important. If one requires an extremely tight confidence interval, then careful stratification and prior construction, and a generous sample, are all recommended.

## Acknowledgments

Los Alamos National Laboratory is operated by the University of California for the United States Department of Energy under contract W-7405-ENG-36. We thank the TIPSTER program and the Linguistic Data Consortium for making the *Federal Register* corpus available, and Mike Cannon and Tony Warnock for helpful discussions on statistical issues.